\documentclass[final]{cvpr}

\usepackage{times}
\usepackage{epsfig}
\usepackage{graphicx}
\usepackage{amsmath}
\usepackage{amssymb}
\usepackage[pagebackref=true,breaklinks=true,colorlinks,bookmarks=false]{hyperref}

\def\cvprPaperID{21} % *** Enter the CVPR Paper ID here
\def\confYear{CVPR 2021}
\begin{document}

%%%%%%%%% TITLE
\title{Sketch-QNet: A Quadruplet ConvNet for Color Sketch-based Image Retrieval}

\author{Anibal Fuentes, Jose M. Saavedra \\
Impresee Inc.\\
600 California St, San Francisco, USA\\
{\tt\small \{anibal.fuentes, jose.saavedra\}@impresee.com}
}

\maketitle

\begin{abstract}
   Architectures based on siamese networks with triplet loss have shown outstanding performance on the image-based similarity search problem. This approach attempts to discriminate between positive (relevant) and negative (irrelevant) items. However, it undergoes a critical weakness. Given a query, it cannot discriminate weakly relevant items, for instance, items of the same type but different color or texture as the given query, which could be a serious limitation for many real-world search applications. Therefore, in this work, we present a quadruplet-based architecture that overcomes the aforementioned weakness. Moreover, we present an instance of this quadruplet network, which we call \textbf{Sketch-QNet}, to deal with the color sketch-based image retrieval (CSBIR) problem, achieving new state-of-the-art results. 
   %We also propose another instance of the quadruplet network that approaches the clothing retrieval problem to show the generalization of our proposal.
   \newline
\newline
\copyright 2021 IEEE
\end{abstract}

%%%%%%%%% BODY TEXT
\section{Introduction}
\begin{figure}[ht]
    \centering
    \includegraphics[width=6cm]{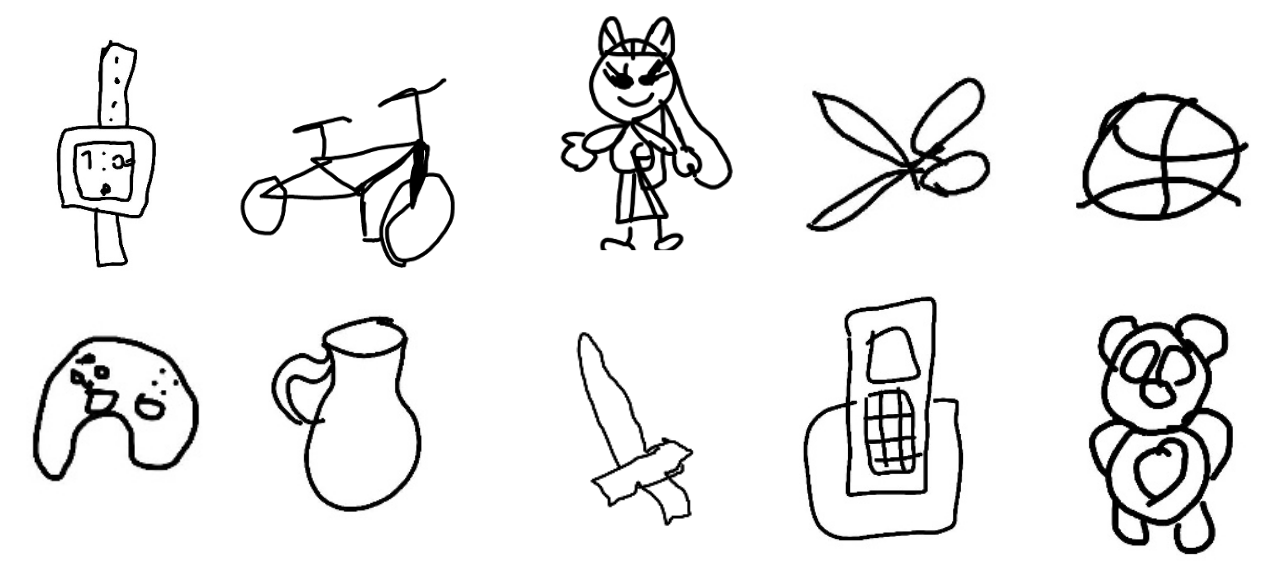}
    \caption{Examples of simple sketches}
    \label{fig:sketches}
\end{figure}

Sketch-based image retrieval (SBIR) is a growing field in computer vision that consists of retrieving a collection of photos resembling a sketched query. Aiming to make the querying process as easy as possible, the input query is formulated as a simple hand-drawing, composed uniquely of strokes as showed in Fig. \ref{fig:sketches}. In this way,  people just need to draw what they are looking for or what they are thinking of.

An SBIR system brings tremendous advantages in different searching contexts. For instance, in e-commerce, the search process is a critical component since it is the first contact with a potential buyer. Thus, Thus, the success of e-commerce will depend on the success of its search engine. In a world where customer engagement seems to be a key concept, SBIR plays an important role, not only in terms of effectiveness but also in terms of customer experience. Moreover, the widespread use of mobile devices fosters this kind of querying modality, since touch-screen interfaces allow people to draw easily and in an entertaining way.

During the last decade, we have seen a resurgence in the interest in SBIR. Works based on low-level features \cite{Saavedra:2014,Eitz:2011,Hu:2013} and mid-level representations \cite{Saavedra:2015} marked the beginning of this resurgence period. However, the explosion of deep learning-based approaches also brings significant improvements to SBIR models \cite{Yu:2016, Sangkloy:2016}. Furthermore, some researchers were also focused on optimizing the deep visual representations through deep hashing-based approaches \cite{Liu:2017}. After all, the architectures based on siamese backbones with triplet loss trained incrementally showed the best performance on different SBIR datasets \cite{Bui:2018}. More recently, a sketch representation generated from a transformer-based model was proposed by Ribeiro et al. \cite{Ribeiro:2020} with competitive results.

Although sketch-based retrieval has shown to be a powerful tool for searching environments like e-commerce, it also has some limitations related to its expressiveness capacity. The input sketch is composed of strokes that represent the shape of the product that the user is looking for, but it lacks relevant attributes in visual perception like color or textures, which are also relevant in applications like e-commerce search.% For instance, for clothing retrieval, these kinds of visual attributes are critical.

\begin{figure}[ht]
    \centering
    \includegraphics[width=7cm]{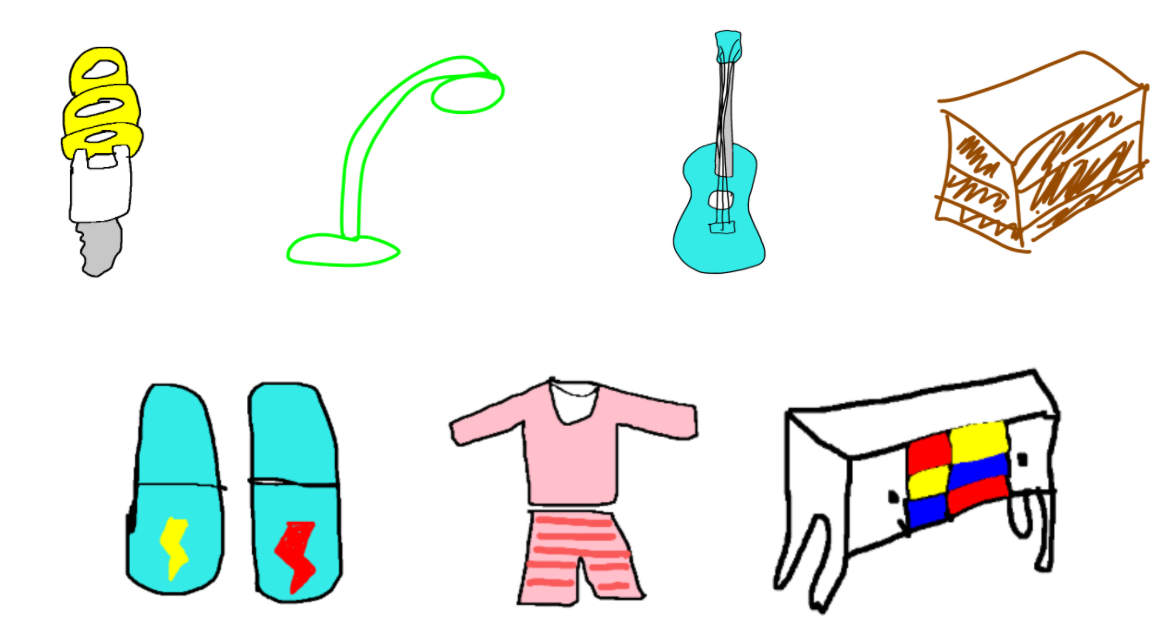}
    \caption{Examples of color sketches}
    \label{fig:color_sketches}
\end{figure}

Contrary to the advances on sketch-based querying, methods dealing with sketches, including color, like those shown in Fig. \ref{fig:color_sketches}, have not followed the same direction, having still few works facing this challenging problem \cite{bui2015scalable},\cite{Xia:2019}. Therefore, our contribution in this work is to propose a quadruplet-based convolutional network architecture that allows us to incorporate an extra visual attribute to the learned feature space. We present an instance of this quadruplet architecture called \textbf{Sketch-QNet} focused on the color sketch-based image retrieval problem. We also describe a methodology for training the proposed architectures because of the scarcity of appropriate datasets.

This document is organized as follows: Section \ref{sec:related_work} describes the related work on CSBIR, Section \ref{sec:quattronet} describes the proposed Sketch-QNet architecture, Section \ref{sec:experimental_evaluation} shows a description of our experiments together with the corresponding results, and finally, Section \ref{sec:conclusions} presents the conclusions of our work.

\section{Related Work}
\label{sec:related_work}
Sketch-based image retrieval has undergone significant growth during the last decade. Initially, through low-level-based approaches  \cite{Saavedra:2012, Hu:2013, Eitz:2011} that use oriented gradients as features. Later, with mid-level-based approaches like the work proposed by Saavedra et al.\cite{Saavedra:2015}, which combines low-level features with a clustering strategy to learn a set of primitives called \emph{keyshapes}, that are then used to represent sketches.  However, the big leap in performance came with the explosion of the deep learning-based methods. In this context, the architectures showing the best results on similarity search are those combining siamese nets with triplet loss \cite{Yu:2016, Bui:2018, Sangkloy:2016}, especially when fine-grained search is desired. Other researchers have also studied architectures focused on obtaining optimized representations through a binary feature space \cite{Liu:2017}.

Among the diversity of SBIR methods, the work of Bui et al. \cite{Bui:2018} mainly attracts our interest. This method proposes a 4-stage incremental methodology for training a network capable of producing a feature space where sketches and photos can exist together. The four stages are designed in such a way that they can incrementally improve their discriminatory power. To this end, they also use siamese and triplet networks jointly with cross-entropy loss but trained from a coarse-grained similarity at the beginning to a fine-grained similarity at the end. This idea showed to be very effective, as it achieves the state of the art results in different public SBIR datasets with colorless sketches.

In color sketch-based image retrieval, the goal is to retrieve photos sharing both shape and color w.r.t. a given query, in such a way that it allows us to formulate queries for retrieving targets like \emph{that special red-blue chair} or \emph{a fancy blue lamp}. Although we have seen tremendous advances in the direction of sketch-based retrieval, there are still few works dealing with colorful sketches. A recent approach dealing with color sketches is the method proposed by  Dutta and Biswas \cite{style-sbir:2020}, where a colorless sketch is used to define the shape of the target images, and color information is provided by a keyword. However, in the domain of querying by sketches, we can incorporate visual attributes into the drawing itself, making the query formulation easier for users. 

Closer to our proposal is the work presented by Bui and Collomose  \cite{bui2015scalable}, which processes shape and color independently using the well-known BoW strategy and features produced by gradient fields. Similar to this work is that proposed by Xia et al. \cite{Xia:2019} which extracts shape features through a convolutional neural network to retrieve the $N$ most similar images. The resulting ranking is then resorted using color histograms. In the two proposals, two scores are computed, one for shape similarity and the other for color. Both scores are merged at the end, setting a weight for each attribute. Contrary to these approaches, we aim to construct a convolutional architecture capable of representing shape and color information in one model. %That means, at the end our model will produce only one feature vector for an input image, carrying shape and color information.

So far, we know that siamese architectures alongside a triplet loss have shown the best performance in the context of similarity search. These architectures are designed to receive triplets as inputs. A triplet is defined as a ($I_a$, $I_p$, $I_n$) tuple, where $I_a$ is an anchor image, the sketch in our case. $I_p$ is a positive photo that is considered similar to the query, while $I_n$ is a negative photo, one considered different from the sketch. Therefore, under this approach, the model tries to learn a feature space where positive pairs come closer to each other, while negative pairs remain afar. Indeed, triplet loss tries to minimize the following expression:
\begin{equation}
\label{eq:triplet_loss}
\max(0, \; \delta(I_a, I_p) + \lambda - \delta(I_a, I_n))
\end{equation}
where $\delta(\cdot)$ is a distance function, and $\lambda$ is the minimum separation between positive and negative distances (the margin).

As we can observe from Eq. \ref{eq:triplet_loss}, given a query $q$, the triplet loss tries to separate positive photos from negative ones by a margin of $\lambda$, no matter how different or similar the positive pairs are from $q$. When we incorporate other attributes like color, we would expect the feature space to be able to differentiate between photos sharing shape and color from those that only share the shape. Designing an architecture capable of incorporating this kind of extra visual information is a challenge we need to deal with.

Therefore, in this work, we propose a novel quadruplet architecture named Sketch-QNet for learning a feature space in the context of color sketch-based image retrieval where colorful sketches and photos can exist together. Our work is inspired by siamese networks as well as the incremental methodology proposed by Bui et al. \cite{Bui:2018}. 

Close to our proposal is the work of Seddati et al. \cite{Seddati} about quadruplets networks for SBIR. However, there are significant differences.  They use a quadruplet architecture in the context of colorless sketch-based retrieval to discriminate between two kinds of positive objects, those sharing the same class and those representing the same instance for an input query. In contrast, our proposal focuses on color sketch-based retrieval. Our model uses a cross-entropy branch to allow the model to retrieve objects from the same class as the query.  Furthermore, the architecture we propose is focused on discriminating objects sharing color and shape with the query (\emph{the positives}) from those that only share shape (\emph{the positive-negatives}). Consequently, our approach requires defining a suitable methodology to form up quadruplets that serve as inputs to the proposed model.

\section{Sketch-QNet for CSBIR}
\label{sec:quattronet}
In this section, we describe \textbf{Sketch-QNet}, a quadruplet convnet for CSBIR able to produce a shared feature space for photos and color sketches that takes into account shape and color features. Inspired from \cite{Bui:2018}, we propose a partially shared network and a training methodology in multiple stages. The basis of our proposal is for the network to learn shape-based features in the early stages of the training and jointly learn shape and color features in the last stage. This is achieved by using a quadruplet neural network with a custom loss function.

%%%%%%%%%%%%%%%%%%%%%%%%%%%%%%%%%%%%%%%%%%%%%%%%%%%%%%%
\subsection{Idea Behind The Method}\label{idea}

\begin{figure}
    \centering
    \includegraphics[width=7cm]{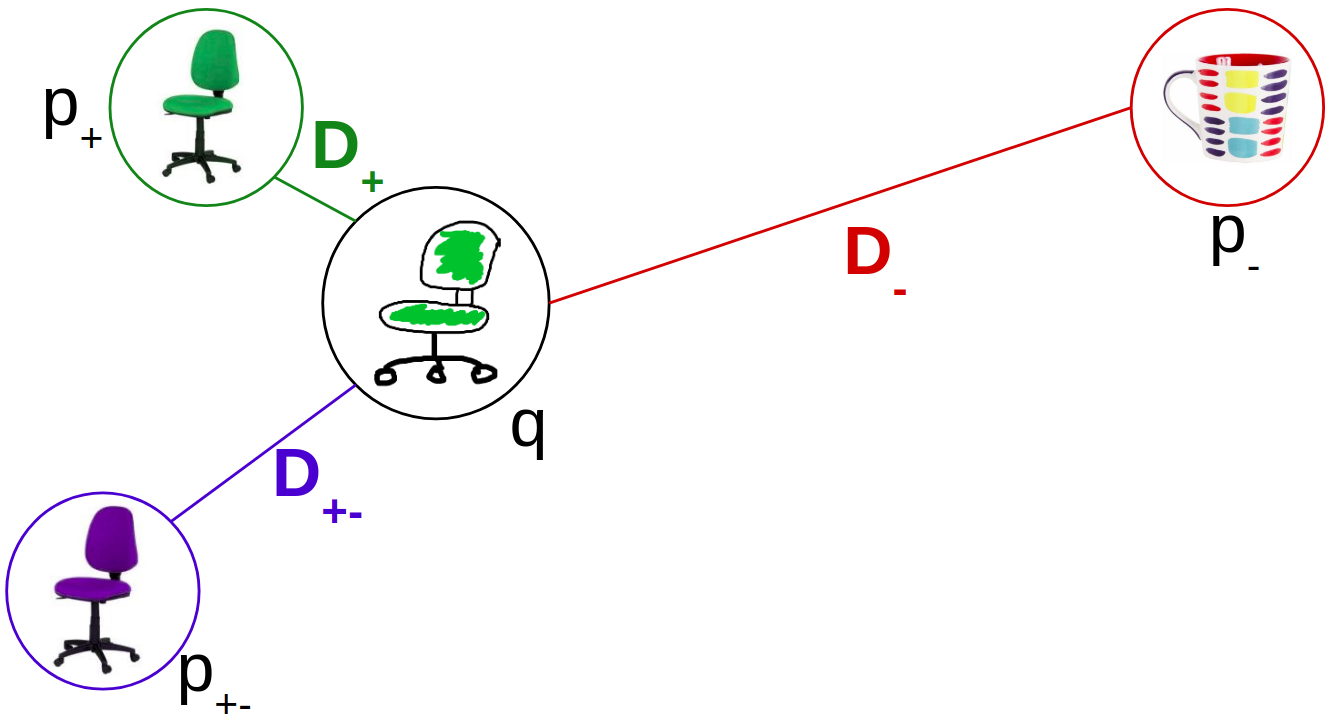}
    \caption{Distances between images must comply  $D_{+} < D_{+-} < D_{-}$.}
    \label{fig:distances_example}
\end{figure}

The core of our proposal is the use of a \emph{quadruplet}, which will be used to feed our model during training. A quadruplet $\Gamma_i$ is defined as follows:
\begin{equation}
\label{eq:quadruplet}
    \Gamma_i = (q^i, p_{+}^i, p_{+-}^i, p_{-}^i)
\end{equation}
where, $i$ is the $i$-th input during the training phase,  $q$ is a color sketch, $p_{+}$ is a photo with the same class and color as $q$, $p_{+-}$  is a photo of the same instance than $p_{+}$ but with a different color, and finally ${p_{-}}$ is a photo from a different class than $q$. A quadruplet must be accompanied by a distance function $D(\cdot,\cdot)$, applied as follows:
\begin{align}
    D_{+} &= D(q,p_{+}) \\
    D_{+-}&= D(q,p_{+-}) \\
    D_{-} &= D(q,p_{-})
\end{align}
without loss of generality, we define $D(\cdot, \cdot)$ to be the Euclidean distance. Our target goal is to train a network capable of producing a feature space where the following relations must be accomplished:
\begin{equation}\label{inequality_1}
    D_{+} < D_{+-} < D_{-}
\end{equation}
A schematic illustration of how the target feature space should behave is depicted in Fig. \ref{fig:distances_example}.

Therefore, we need to appropriately train a convnet for extracting embeddings from photos and sketches in order to fulfill the inequalities in Eq. \ref{inequality_1}. To solve this problem we separate the equation into two inequalities and add a margin to each one, obtaining Eq. \ref{inequality_2}.

\begin{equation}\label{inequality_2}
    \begin{split}
        D_{+} + \alpha \cdot \lambda & < D_{+-} \\
        D_{+-} + (1-\alpha)\cdot \lambda & < D_{-}
    \end{split}
\end{equation}

In Eq. \ref{inequality_2}, $\lambda$ is a positive number representing the margin between  $D_{+}$ and $D_{-}$, and $\alpha$ is a trade-off parameter between 0 and 1 which indicates how  closer is $p_{+-}$ from $p_{+}$ and $p_{-}$. In this work, $p_{+-}$ is considered as a positive sample, but not as relevant as $p_{+}$. Therefore we choose $\alpha < 0.5$ to allow $p_{+-}$ to get closer to $p_{+}$ than to $p_{-}$. We will evaluate the impact of $\alpha$ afterwards.

The  following two losses are inferred from Eq. \ref{inequality_2}:
%The two relations showed in Eq. \ref{inequality_2} yield the following two triplet loss functions  :

\begin{equation}\label{loss_functions}
    \begin{split}
        Loss_{triplet\;1} &= \max\{0, D_{+} + \alpha \cdot \lambda - D_{+-}\} \\
        Loss_{triplet\;2} &= \max\{0, D_{+-} + (1-\alpha) \cdot \lambda - D_{-}\}
    \end{split}
\end{equation}

%hich represents two triplet losses, the %irst one between  $q$, $p_{+}$, and $p_{+ -}$; and he second one between $q$, $p_{+ -}$, and $p_{-}$. So, the network must be trained using this two losses simultaneously and, consequently, having four branches, one for the sketch and the other three for the photos.
It is interesting to note the dynamic of the proposed loss functions during training. At the beginning, neither $Loss_{triplet\;1}$ nor  $Loss_{triplet\;2}$ are less than $0$. Therefore the optimizer will focus on minimizing the expression $D_{+} + \lambda - D_{-}$. It will try to separate positives from negatives samples according to the margin $\lambda$. Accomplished that, the optimizer will focus on simultaneously satisfying the two specific losses of Eq. \ref{loss_functions}.

%In practice, during the first iterations of training, the network  will try to accomplish $D_{-} - D_{+} > \lambda$, then it will try to complain with $D_{+ -} - D_{+} >  \alpha \cdot \lambda$.

\subsection{Training Quadruplets and Dataset}\label{sec:dataset}
In this section, we describe how quadruplets are built during training in order to optimize our convolutional architecture according to the proposed loss function (Eq. \ref{loss_functions}). 

As shown in Eq. \ref{eq:quadruplet}, for each $q_i$, we need to select three regular images (photos) that complete the quadruplet $\Gamma_i$. However, due to the lack of a suitable dataset focused on this problem, we develop a methodology for training our proposed architecture where color queries are generated from a target photo. 

\begin{figure}
    \centering
    \includegraphics[width=7cm]{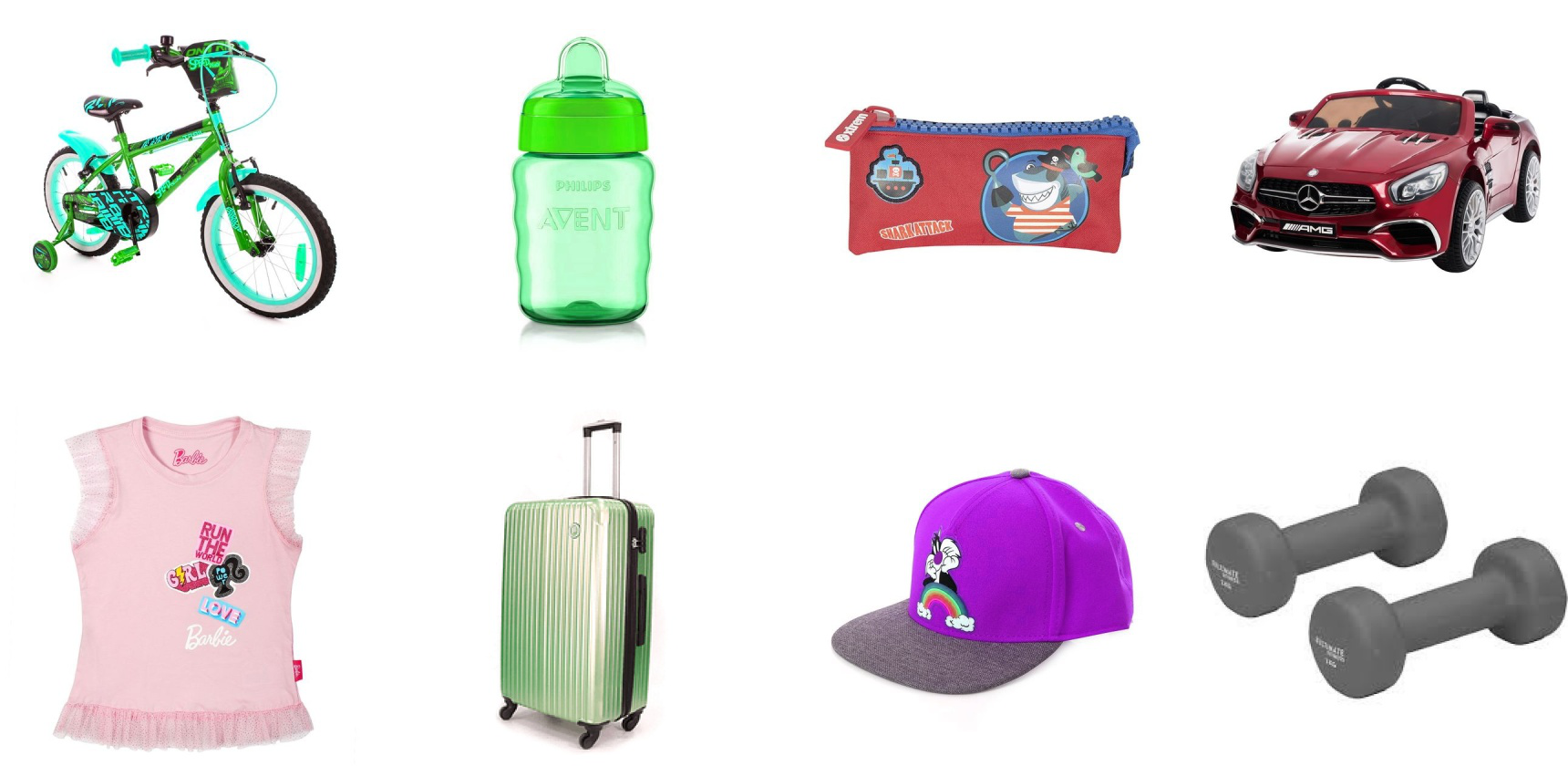}
    \caption{Example of training photos.}
    \label{fig:datasets_examples}
\end{figure}

For training, we use a collection of photos obtained from an online store. This dataset consists of 10,600 items distributed among 140 classes, mainly formed with kids and household items. Some of the products we can find in this collection belong to categories like toys, clothing, baby items, appliances, sports, among others. Figure \ref{fig:datasets_examples} shows a sample of the training photos.

Since the system must be able to distinguish color, we increase the number of items in the training collection by changing the colors from the original images. To this end, each item is processed in the HSV color space, adding a constant value to the hue channel. This allows the dataset to grow by a factor of 5. We also generate gray-toned photos of different brightness, transforming the original images to gray-scale, segmenting the foreground by thresholding, and applying a contrast enhancement. The final extended dataset consists of 56,337 items.

%We extended the training collection by changing the hue value of each item in the HSV color space, yielding 50482 instances. For each image from this collection we generate a synthetic sketch, 

\begin{figure}
    \centering
    \includegraphics[width=8cm]{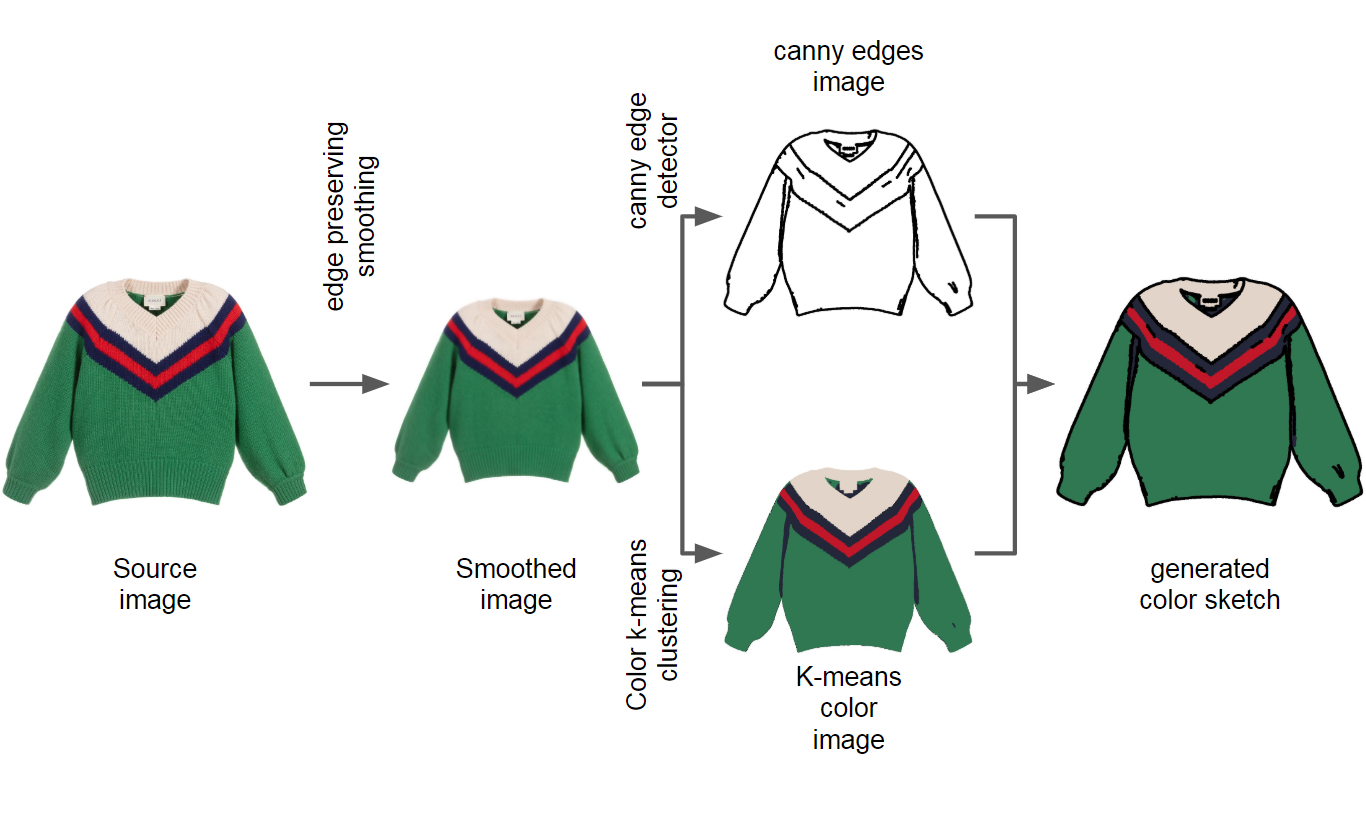}
    \caption{Color sketch generation process.}
    \label{fig:sketches_generations}
\end{figure}

A critical problem when training a neural network that learns color sketch-based embeddings is the lack of datasets with plenty of color drawings. To handle this problem, we generate a synthetic color sketch from each item of the training catalog through a function $s(p)$, where $s(\cdot)$  transforms the photo $p$ to its corresponding sketch. This process is illustrated in Fig. \ref{fig:sketches_generations}. First, $p$ is smoothed by an edge-preserving smoothing filter \cite{edgepreservingfilter}, then its color range variation is reduced to have flat colors, and finally, edges are highlighted. To obtain flat colors, we apply k-means clustering in the RGB space over the smoothed images. The number of clusters is selected randomly between 7 and 10. To highlight edges, we apply the Canny edge detector \cite{canny} over the smoothed image. 

Therefore, a quadruplet $\Gamma_i$ as defined in Eq. \ref{eq:quadruplet} is generated  by each item $p_i$ from the training collection. Here the anchor sketch $q$ is $s(p_i)$, $p_{+}$ is $p_i$, $p_{+-}$ is $p_i$ but with a different color, finally, $p_{-}$ is randomly selected from items having a different class than $p_i$.

\subsection{Backbone Architecture}

We use a Squeeze and Excitation \cite{Squeeze_and_excitation} ResNext 50 \cite{resnext} network as backbone. Since this is a cross-domain problem including photos and color sketches, we need two network branches to produce the embeddings. The first layers of both networks use non-shared weights to extract specific features from each domain, while the final layers share weights to learn a common representation between both domains. 

\begin{figure*}
    \centering
    \includegraphics[width=16cm]{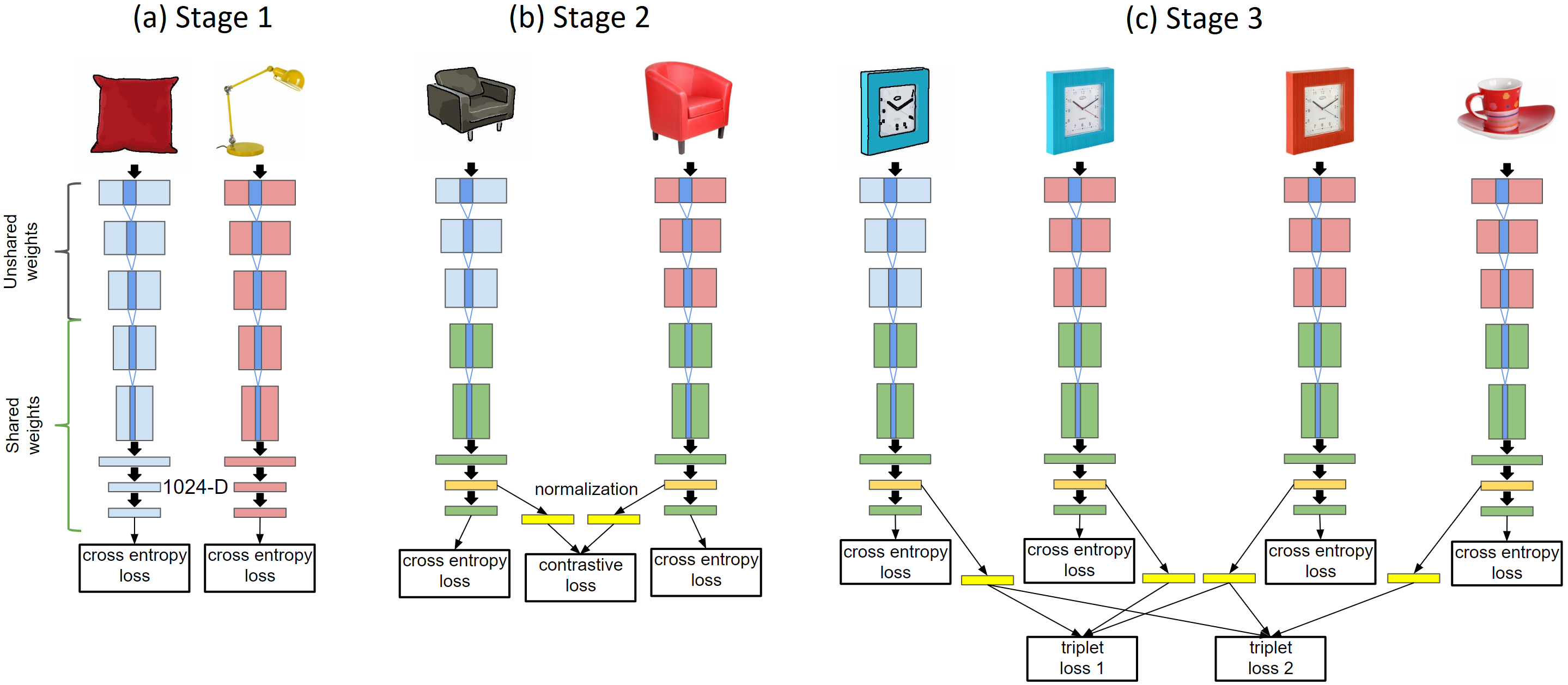}
    \caption{Architecture and training stages.}
    \label{fig:training_procedure}
\end{figure*}

At the top of the network, two fully connected layers are used. The last one is devoted to learn a 1024-d feature vector that is then normalized by the Euclidean norm. This layer also feeds a classification layer. The complete Sketch-QNet architecture is presented in Fig. \ref{fig:training_procedure}.

\subsection{Training Procedure}\label{sec:training_procedure}

The training procedure is inspired by the work of Bui et al. \cite{Bui:2018} in which an incremental training procedure is carried out. In our case, we divide the procedure into three stages, as shown in Fig. \ref{fig:training_procedure}, training each one as follows:
\begin{itemize}
 \item \textbf{Stage 1}: This stage aims to train the backbones before they are plugged into the quadruplet architecture. In this step, each backbone is trained in a classification problem with the classical cross-entropy loss function. The sketch backbone is trained using the sketch dataset proposed by Eitz et al. \cite{Eitz:2011}, and the photo backbone is trained on the ImageNet dataset \cite{imagenet}.

  %considering the problem as a classification problem, where the objective function to minimize corresponds to cross-entropy loss of the network output and real classes.

\item \textbf{Stage 2}: This stage aims to train both networks together, trying to learn a shared embedding for both, photos and sketches. The loss is defined as a multitask function that combines contrastive loss with cross-entropy. The network is trained so that the learned embeddings produce short distances between objects from the same class and a large distance between elements from different classes. The definition of the used loss is shown in Eq.  \ref{objective_2}, where $\beta$ is set to 2 during training, $\text{CE}_{sk}$ is the cross-entropy for sketch classification, and $\text{CE}_{ph}$ is for photo classification.
\begin{align}
\label{objective_2}
    loss_{pt\;2} = & \text{CE}_{x_{sk}}+\text{CE}_{x_{ph}}+  \\ 
    &  \beta \cdot contrastive(x_{sk},\; x_{ph}) \nonumber
\end{align}
\item \textbf{Stage 3}: The goal of this stage is to learn embeddings that take into account color and shape features, according to Section \ref{idea}. Thus, a quadruplet network is trained receiving four inputs, the first one is an anchor sketch and the rest are photos. In this way, we have four branches, the first one processing the input sketch and the other three focused on processing photos.

 The objective function to minimize is presented in Eq. \ref{objective_3}. It consists of adding the cross-entropy loss to each branch to provide information about the classes, together with the triplet losses given in Eq. \ref{loss_functions}. The weight $\beta$ is initialized at 2, and increased by 0.5 each epoch to give more importance to triplet losses as training progresses.

\begin{equation}
\label{objective_3}
    \begin{split}
    loss_{pt\;3} = & \text{CE}_{x_{sk}}+\text{CE}_{x_{p+}}+\text{CE}_{x_{p+-}}+\text{CE}_{x_{p-}}+ \\
    & \beta \cdot \{ Loss_{triplet\;1} + Loss_{triplet\;2} \}
    \end{split}
\end{equation}
\end{itemize}

\section{Experimental Results}\label{sec:experimental_evaluation}
\subsection{Experimental Settings}
We use the following experimental settings for training:

\begin{itemize}
    \item In stages 1 and 2, the network was trained for 10 epochs, while in stage 3 it was trained for 25 epochs, and the best epoch was selected by evaluating the loss on the validation set.
    
    \item The sketches are centered, and white padding of about $10\%$ of the largest dimension is added. In the case of photos, they are resized, keeping their aspect ratio, and adding a white background if necessary. Both sketches and photos are resized to  $224 \times 224$ before passing through the network. 
    \item Data augmentation was applied to both photos and sketches, consisting of random rotation between $-20$ and $20$ degrees, random resize between $90\%$ and $110\%$ of the original size of the images, random flip and color augmentation, changing contrast, saturation and the value of the hue channel.
\end{itemize}

\subsection{Baseline Approach}

So far we know, the current methods dealing with color sketches for image retrieval are the approaches proposed by Bui and Collomosse \cite{bui2015scalable} using gradient fields together with a BoW strategy; and the proposal of  Xia et al. \cite{Xia:2019} that uses a convnet to extract shape features. In general terms, both proposals are similar as they have two independent paths to process shape and color, producing two scores for each of these attributes. In the end, these scores are weightily merged to compute the final one. 

Based on the specifications provided by the aforementioned CBIR proposals, we implemented two baselines that are described below. 

\subsection{Baseline 1}
For shape description, we trained a convnet following the work of Bui et al. \cite{Bui:2018}, using a SE-ResNext-50 \cite{Squeeze_and_excitation} as backbone. For color description, we use a color histogram computed from RGB color spaces,  quantized into $5\times5\times5$ cells. In addition, to keep spatial coherence, we split the images into a $2\times2$ grid, computing the color histogram for each cell, and finally, all of them are concatenated. Therefore, this baseline produces a 500-size color histogram.

For similarity search, we first compute two distances, one for shape and the other for color description that then are merged by a  weighted distance function as shown in Eq. \ref{baseline_distance}. Here,  $D(S_{\text{sk}}, S_{\text{ph}})$ is the distance between shape representations of a sketch and a photo, $D(C_{\text{sk}}, C_{\text{ph}})$ is the distance between the corresponding color based representations, and $\gamma$ is a weight between 0 and 1. 

%The latter vector was computed using a 3D color vector quantization \cite{3d_color_vector_quantization}, which consists in discretizing the RGB color space into a 5x5x5 grid and the image in a 2x2 grid, and building a color histogram for the 4 cells of the image, obtaining a color feature vector of dimensionality 500. Unkike \cite{Xia:2019}, to apply similarity search, we formulate a weighted distance function between the distance produced by shape and color as shown in equation \ref{baseline_distance}. Here,  $D(S_{\text{sk}}, S_{\text{ph}})$ is the distance between shape representations of a sketch and a photo, $D(C_{\text{sk}}, C_{\text{ph}})$ is the distance between the corresponding color based representations, and $\gamma$ is a weight between 0 and 1.

\begin{equation}
\label{baseline_distance}
    D(\text{sk},\text{ph}) = (1-\gamma) \cdot D(S_{\text{sk}}, S_{\text{ph}}) + \gamma \cdot D(C_{\text{sk}}, C_{\text{ph}})
\end{equation}

\subsubsection{Baseline 2}
This baseline uses the same strategy as Baseline 1 to describe shapes and slightly modifies the color representation. The differences in color processing are that we do not use spatial splitting, and only the pixels of strokes are considered to construct the color histogram. A bigger difference is related to the manner in which similarity is computed. Here, for color-based similarity we use the same function as proposed by Bui and Collomosse\cite{bui2015scalable},  that is defined as below.
%\ref{baseline_2_color_distance}.
%, where  $\text{Sim}_c$ is the color similarity  between a sketch $\text{sk}$ and the $i-th$ photo $\text{ph}_i$ from a catalog.
%The second baseline, inspired by the work of Bui and Collomose \cite{bui2015scalable}, adresses the shape feature extraction in the same way as the first baseline, but changes the way to compute the color features. To calculate this features, a color histogram is extracted by discretizing the RGB space in a 5x5x5 grid, obtaining 125 bins $b$ for each image and sketch; it's important to emphasize that for the sketch only the pixels where the user drew are added to the histogram, and not the background pixels. Then, to calculate the color similarity  between a sketch $\text{sk}$ and a photo $\text{ph}_i$ of the dataset, the expression of eq. \ref{baseline_2_color_distance} is used

\begin{equation}
\begin{split}
\label{baseline_2_color_distance}
    \text{Sim}_c & (\text{sk},\text{ph}_i) = \frac{1}{M_{\text{sk}} M_{\text{ph}_i}} \times \\
    & \sum_{b \in \text{sk} \cap \text{ph}_i} (1+\text{ln}f_{\text{sk},b}) (1+\text{ln}f_{\text{ph}_i,b}) \text{IDF}_b,
\end{split}
\end{equation}

where $M_{\text{sk}}=\sqrt{\sum_{b \in \text{sk}} (1+\text{ln}f_{\text{sk},b})^2 }$, $M_{\text{ph}_i}=\sqrt{\sum_{b \in \text{ph}_i} (1+\text{ln}f_{\text{ph}_i,b})^2}$, $\text{IDF}_b = 1+\text{ln}\frac{N}{f_b}$, $N$ is the number of indexed images in the dataset, $f_b$ is the number of images containing the bin $b$, $f_{\text{sk},b}$ and $f_{\text{ph}_i,b}$ are the counts of the bin $b$ in the query sketch and the photo $i$ respectively. This similarity measure resembles tf-idf for color space. Finally, the similarity between a sketch \text{sk} and a photo $\text{ph}_i$ is presented by Eq. \ref{baseline_2_similarity}, which represents the geometric mean between the color similarity, and the cosine distance of the shape features (shape similarity). 

\begin{equation}
\label{baseline_2_similarity}
\text{Sim}(\text{sk},\text{ph}_i) = \text{Sim}_c  (\text{sk},\text{ph}_i)^\omega (\frac{S_{\text{sk}} \cdot S_{\text{ph}_i}}{||S_{\text{sk}}|| \; ||S_{\text{ph}}||})^{(1-\omega)}
\end{equation}

%This baseline reaches a mAP of $0.30$ in the dataset Flickr15k color \cite{bui2015scalable}, overcoming the results presented in the paper.
Both baselines were tested in the Bui's dataset \cite{bui2015scalable}, where Baseline 2 reaches a mAP of 0.30, outperforming the performance reported previously (mAP=25.8). In contrast, Baseline 1  yielded a lower performance (mAP=24.55) but very close to the state-of-the-art. These baselines will be used for comparison in the following sections.

\subsection{Testing Dataset}
Our interest rests on making the querying formulation easier for users, in such a way that they can simply drawing what they are looking for, incorporating color information in the drawing itself. Therefore, we are interested in testing the proposed methods with queries representing the high variability of drawing styles (see Fig. \ref{fig:color_sketches}). To this end, we present a challenging testing dataset containing sketches that incorporate color in the form of shadows, color stripes, color strokes, etc. To our knowledge, this dataset is the first one that covers such variability of styles, and it is described in the following lines.

The dataset used for testing consists of photos of home \& decor items from online stores, distributed in 187 classes, among kitchen items, garden, decoration, and others. It is worth noting that the evaluation dataset is entirely different from the one used in training. The number of items in the testing catalog is 10560, which was increased to 50482, by using the color augmentation technique presented in section \ref{sec:dataset}. Additionally, we collected 446 hand-drawn sketches based on photos from the dataset.  To this end, we asked ten different people to make drawings resembling some example photos. Each target photo was shown for a few seconds, and then the user drew it from memory.

To evaluate the performance of the described methods, we employ the following metrics, reminding that each sketch has only one groundtruth photo:

\begin{itemize}
\item \textbf{Mean Reciprocal Rank (MRR)} : This measures the average of the inverse rank of the correct answer (groundtruth photo).
\item \textbf{Recall Ratio (RR)} : This shows the percentage of queries that retrieve the groundtruth photo after retrieving $N$ items.
\item \textbf{Mean Average Precision (mAP)} : We use mAP to evaluate the precision of retrieving photos from the same class of the query (regardless of the color).
\end{itemize}
\subsection{Baseline Results}
%The network for deep shape features extraction (SBIR without color) was tested using the \textit{Flickr15K} dataset \cite{Hu:2013}, reaching a mAP of 0.553. 
\subsubsection{Baseline 1}
 Table \ref{tab:baseline_results}  shows the MRR and mAP achieved by the first baseline approach for different values of $\gamma$ in the range $[0, 1]$. We can observe that smaller values of $\gamma$ produce higher mAP, ignoring color information, focusing instead on the class of the objects. On the other hand, as $\gamma$ increases, then the MRR also increases because the retrieved photos with similar color as the query go to the first positions. This happens until a point where both metrics start decreasing because the retrieval system begins to overuse the color features. We select $\gamma=0.6$ as the best value because it reaches a good trade-off between MRR and mAP (0.1511 and 0.1437, respectively).

\begin{table}[ht!]
\begin{center}
\begin{tabular}{|c|c|c|c|c|c|c|}
\hline
\textbf{$\gamma$} & 0.0   & 0.2   & 0.4   & \textbf{0.6}   & 0.8   & 1.0   \\ \hline
\textbf{MRR}      & 0.081 & 0.105 & 0.138 & \textbf{0.151} & 0.143 & 0.022 \\ \hline
\textbf{mAP}      & 0.160 & 0.163 & 0.160 & \textbf{0.143} & 0.095 & 0.011 \\ \hline
\end{tabular}
\end{center}
\caption{MRR and mAP for the baseline 1 changing the value of $\gamma$.}
\label{tab:baseline_results}
\end{table}

\begin{table}[ht!]
\begin{center}
\begin{tabular}{|c|c|c|c|c|c|c|}
\hline
\textbf{$\omega$} & 0.0   & \textbf{0.05}  & 0.1   & 0.2   & 0.4   & 0.6   \\ \hline
\textbf{MRR}      & 0.081 & \textbf{0.150} & 0.150 & 0.118 & 0.057 & 0.030 \\ \hline
\textbf{mAP}      & 0.160 & \textbf{0.125} & 0.099 & 0.057 & 0.024 & 0.013 \\ \hline
\end{tabular}
\end{center}
\caption{MRR and mAP for the baseline 2 changing the value of $\omega$.}
\label{tab:baseline_2_results}
\end{table}

\subsubsection{Baseline 2}
Table \ref{tab:baseline_2_results} shows the MRR and mAP achieved by this approach for different values of $\omega$ in the range $[0, 1]$. As in the first baseline, $\omega$ has a similar effect to $\gamma$, but achieves a lower performance than the first baseline. This is related to the fact that the method does not consider spatial information in the color features.

\subsection{Sketch-QNet Results}

We follow the training procedure explained in section \ref{sec:training_procedure}. We train the network with the loss of Eq. \ref{loss_functions} with $\lambda=1.5$ and $\alpha \in \{0.1, 0.25, 0.5, 0.75\}$. The evaluation metrics from stages 2 and 3 of the training are shown in Table \ref{tab:results_quattronet}. The best mAP is reached at stage 2 of the training process because this stage considers class information only. On the other hand, in stage 3,  mAP decreases, but MRR increases more than $50\%$ compared to stage 2. This growth in MRR is explained by the fact that stage 3 focuses on bringing the photos of the same \textbf{color and shape} as the query closer to it in the feature space. We can also note that higher values of $\alpha$ make mAP decrease because when $\alpha$ is larger, photos of the same class but with a different color than the sketch are mapped further away from it. This effect shows that $\lambda$ in our proposal has a similar behavior as the $\gamma$ parameter in the first baseline approach.

\begin{table}[h]
\begin{center}
\begin{tabular}{|c|c|c|c|c|c|c|}
\hline
\textbf{Stage} & \begin{tabular}[c]{@{}c@{}}Stg.\\ 2\end{tabular} & \begin{tabular}[c]{@{}c@{}}Stg 3\\ $\alpha{=}0.10$\end{tabular} & \begin{tabular}[c]{@{}c@{}}Stg. 3\\ $\alpha{=}0.25$\end{tabular} & \begin{tabular}[c]{@{}c@{}}Stg. 3\\ $\alpha{=}0.50$\end{tabular} & \begin{tabular}[c]{@{}c@{}}Stg. 3\\ $\alpha{=}0.75$\end{tabular} \\ \hline
\textbf{MRR}   & 0.126                                           & \textbf{0.208}                                                & 0.206                                                          & 0.197                                                          & 0.192                                                          \\ \hline
\textbf{mAP}   & 0.186                                            & 0.133                                                         & 0.083                                                          & 0.051                                                          & 0.037                                                          \\ \hline
\end{tabular}
\end{center}
\caption{MRR and mAP for Sketch-QNet at different stages of training.}
\label{tab:results_quattronet}
\end{table}

\begin{figure*}
    \centering
    \includegraphics[scale=0.35]{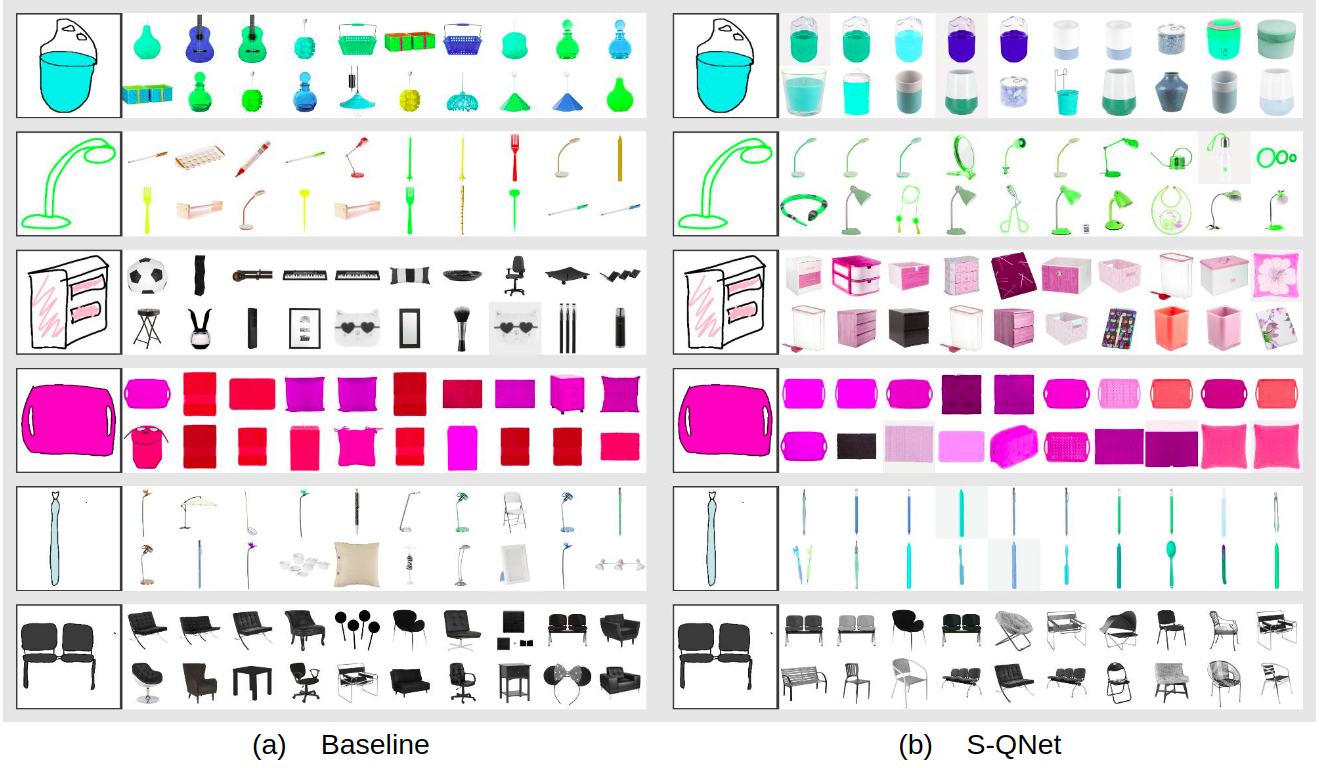}
    \caption{Example of top 20 retrieved photos for baseline 1 (left) and stage 3 of S-QNet (right).}
    \label{fig:retrieved_images}
\end{figure*}

%\begin{figure*}
%    \centering
%    \includegraphics[width=14cm]{images/cluster.jpeg}
%    \caption{Cluster of slippers, visualized through t-SNE using the whole dataset, in red the query sketches.}
%    \label{fig:cluster_sqnet}
%\end{figure*}

\begin{figure}[h]
    \centering
    \includegraphics[width=7cm]{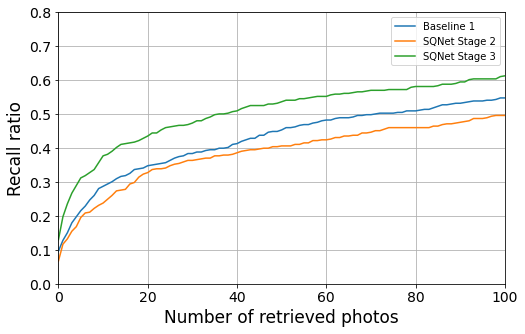}
    \caption{Recall ratio for baseline 1, stage 2 and 3 of Sketch-QNet.}
    \label{fig:recall_ratio}
\end{figure}

Another way to compare the performance of the proposed methods is through a recall ratio chart. In Fig. \ref{fig:recall_ratio} we present the recall ratio for the baseline 1 with $\gamma = 0.6$, the stage 2 of Sketch-QNet and the stage 3 of Sketch-QNet with $\alpha = 0.10$. The figure shows that the proposed method in stage 3 has a higher performance than the baseline. For instance, in stage 3 only 37 photos need to be retrieved (from the 50,482 dataset images) to get a relevant result for $50\%$ of the queries, while for the baseline, 72 images must be retrieved to reach the same percentage. Of course, it is important to note the performance of a method when retrieving a small number of photos, since users commonly prefer to change the query if there is not a hit among the first results.

In Figure \ref{fig:retrieved_images}, we include examples of different retrieved photos, for the baseline 1 and for stage 3. %Furthermore, for a better understanding of the network behaviour, in Fig. \ref{fig:cluster_sqnet} we depict a t-SNE based visualization for the testing dataset, where similar colors tend to be grouped in similar locations.

\section{Conclusions}\label{sec:conclusions}

In this work, we present a novel quadruplet-based convnet architecture together with a loss function that extends the behavior of siamese and triplets networks in the context of the similarity search problem, which is limited to handle positive and negative pairs only. Our proposal allows us to incorporate new intermediate pairs, which can be selected according to visual similarity with the anchor. Therefore, this proposal allows us to generate a feature space that can additionally discriminate among positive results, adding an extra visual feature to the model like color or texture.

To show the advantages of our proposal, we trained an instance of the quadruplet model named Sketch-QNet, that achieves new state-of-the-art results on the problem of color sketch-based retrieval. In addition, to handle the lack of datasets in this problem, we also present a methodology for generating a training dataset from a collection of photos.
%To show how our general proposal can be applied to a different problem, we apply the quadruplet proposal for clothing retrieval, where users are usually interested in getting photos resembling not only the shape, but also color and texture of an input query. We show qualitative results produced by our proposal. 

%The proposed methodology shows to be effective in problems like color sketch based image retrieval and image based clothing retrieval, especially in cases where fine-grained details are needed, and extends the reach of siamese and triplet networks, which are limited by positive and negative pairs, allowing the incorporation of new intermediate pairs, which can be selected according to visual similarity with the anchor, as well as in our case, in which they were selected by their color.

%proposed Sketch-QNet, a method for CSBIR consisting on training a neural network capable of extract a common embedding between photos and sketches, representing shape and color information of the inputs. This method uses a training strategy based on quadruplet networks, with a custom loss function whose objective is to arrange the images according to shape and color information, bringing closer pairs of the same shape and color, pushing away pairs of the same shape but different color, and pushing away further pairs of different shapes. We also proposed a dataset to test the proposed models, consisting of about 50,000 photographs and 446 hand drawn sketches made by drawers with different skills.

{\small
\bibliographystyle{ieee_fullname}
\bibliography{egbib}
}

\end{document}